\documentclass[10pt,twocolumn,letterpaper]{article}

\usepackage[pagenumbers]{cvpr} 
\newcommand{\submissiontype}{review}  

\ExplSyntaxOn
\NewDocumentCommand{\multiversion}{m m m}{
    \str_case:nn {\submissiontype} {
        {review} {#1}
        {arxiv}  {#2}
        {final}  {#3}
        {default}{\textbf{Error: Invalid value for \textbackslash submissiontype. Please set it to "review", "arxiv", or "final".}}
    }
}
\ExplSyntaxOff

\definecolor{cvprblue}{rgb}{0.21,0.49,0.74}
\usepackage[pagebackref,breaklinks,colorlinks,allcolors=cvprblue]{hyperref}
\usepackage{multirow}
\usepackage{footmisc} 

\def\paperID{13002} 
\def\confName{CVPR}
\def\confYear{2025}

\title{ InfinityDrive: Breaking Time Limits in Driving World Models}

\author{
Xi Guo$^{1}$\thanks{Equal contributions.}\hspace{0.6em}
Chenjing Ding$^{1}$\footnotemark[1] \thanks{Group Leader, Email: \texttt{dingchenjing@sensetime.com}}\hspace{1em}
Haoxuan Dou$^1$\hspace{0.6em}
Xin Zhang$^2$\hspace{0.6em}
Weixuan Tang$^1$\hspace{0.6em}
Wei Wu$^{1,2}$\thanks{Corresponding Author, Email: \texttt{wuwei@senseauto.com}}
\\
{\normalsize $^1$SenseAuto \enspace
 $^2$Tsinghua University}\\
{\tt\small \{guoxi,dingchenjing\}@sensetime.com}, {\tt\small xin-zhan22@mails.tsinghua.edu.cn} \\
{\tt\small \{tangweixuan, wuwei, douhaoxuan\}@senseauto.com} \\
}

\begin{document}
\maketitle

\begin{abstract}
Autonomous driving systems struggle with complex scenarios due to limited access to diverse, extensive, and out-of-distribution driving data which are critical for safe navigation. World models offer a promising solution to this challenge; however, current driving world models are constrained by short time windows and limited scenario diversity. To bridge this gap, we introduce InfinityDrive, the first driving world model with exceptional generalization capabilities, delivering state-of-the-art performance in high fidelity, consistency, and diversity with minute-scale video generation. InfinityDrive introduces an efficient spatio-temporal co-modeling module paired with an extended temporal training strategy, enabling high-resolution (576$\times$1024) video generation with consistent spatial and temporal coherence. By incorporating memory injection and retention mechanisms alongside an adaptive memory curve loss to minimize cumulative errors, achieving consistent video generation lasting over 1500 frames (more than 2 minutes). Comprehensive experiments in multiple datasets validate InfinityDrive's ability to generate complex and varied scenarios, highlighting its potential as a next-generation driving world model built for the evolving demands of autonomous driving. 
Our project homepage: \url{https://metadrivescape.github.io/papers_project/InfinityDrive/page.html}

\end{abstract}    
 \section{Introduction}
\label{sec:intro}
\begin{figure*}[ht]
  \centering
   \includegraphics[width=1.00\linewidth]{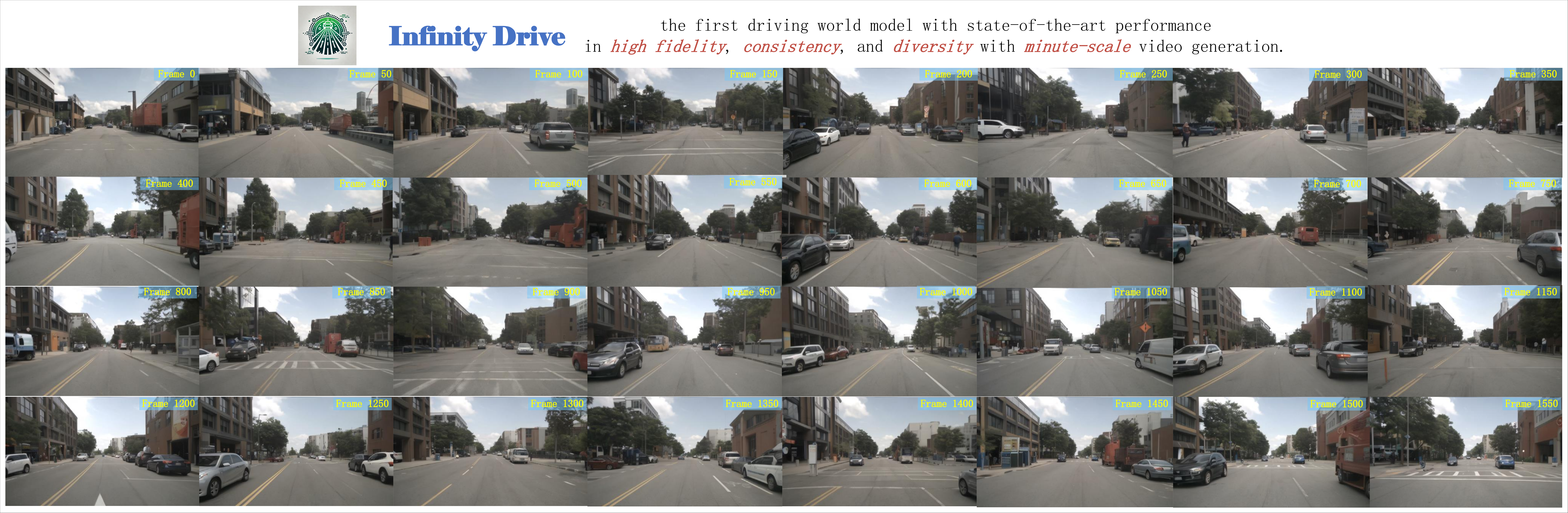}
   \caption{InfinityDrive can generate long-term driving videos up to 1500 frames.}
   \label{fig:longterm_ours}
\end{figure*}

Autonomous driving systems face significant challenges due to limited access to diverse, out-of-distribution data that is essential for safe navigation. World models tackle this by generating high-quality, consistent, and varied long-duration videos that simulate complex, real-world scenarios. Long-term video generation is crucial, as it enables models to capture extended temporal dependencies, allowing vehicles to anticipate, plan, and respond proactively to future events with sufficient reaction time. Additionally, high-resolution video preserves details, and temporal consistency ensures realism. The diversity of simulated scenarios further enhances adaptability, enabling the system to handle a wide variety of conditions. Together, these capabilities equip autonomous systems to make accurate, long-term decisions in dynamic environments.

However, existing driving world models still face significant limitations when tasked with producing long-duration video sequences(as shown in Fig.~\ref{fig:longterm}): \textbf{1)}Poor Spatio-Temporal Resolution: Many current driving world models~\cite{vista, svd, drivedreamer,wovogen,adriver-i,drivedreamer2,genad}, operate within a relatively short temporal window ($<$30 frames), resulting in generated sequences lasting only a few seconds. This constraint is due to the models' limited ability to encode and process features over extended time frames. Models~\cite{gaia-1, streamingT2V, loong} which attempt long-term generation, often resort to lower resolutions, sacrificing crucial details needed to represent complex driving environments accurately. This limitation stems from the significant GPU memory requirements associated with high-resolution spatio-temporal modeling, leading to a reduction in the quality and realism of simulated driving scenarios. \textbf{2)}Autoregressive Error Accumulation: Driving world models~\cite{vista, svd, streamingT2V} perform long-term rollouts by iteratively predicting short-term clips and resetting the condition image with the last clip. However, prediction inaccuracies cause small deviations from the original conditions, which amplify over time and lead to significant drift from the true state in long sequences, reducing accuracy and consistency of generated videos. \textbf{3)}Lack of Diversity: Current models~\cite{vista, svd, drivedreamer,wovogen,drivedreamer2,genad} exhibit limited diversity in their outputs, producing nearly identical videos when conditioned on the same initial input and random noise. This limitation hinders the model's ability to generate diverse driving situations.

To address the limitations of existing world models in autonomous driving, we propose a series of innovations that enhance both the quality and diversity of long-term video generation. We compare the characteristics of our method with other methods and summarize them in Tab~\ref{tab:comparision of world models}. Our model demonstrates superior capability in spatio-temporal resolution and long-horizon consistent rollout.

\textbf{Firstly}, to tackle the challenge of spatio-temporal resolution, we introduce the Efficient Spatio-Temporal Co-Modeling module, which dynamically adjusts the information density based on the resolution, prioritizing spatial detail at higher resolutions while enhancing temporal modeling at lower resolutions. This module ensures high fidelity and consistency in long-horizon video generation.
In addition,  we introduce an extended temporal scope training technique using a curriculum learning strategy that progressively expands the temporal window to 128 frames per iteration. 
This approach enables the model to predict further into the future and effectively model these long-term dependencies and behavioral trends. 
\textbf{Secondly}, to solve the error accumulation problem of long-duration predictions, we introduce Memory Injection and Retention Mechanisms, which ensure that the model retains accurate memory throughout long-term inference, preventing errors from accumulating over time. In addition, inspired by the Ebbinghaus forgetting curve~\cite{ebbinghaus2018gedächtnis}, we propose a memory curve adaptive loss function to enhance memory retention for near-future frames, thereby preventing error propagation, while simultaneously fostering generative flexibility for distant-future frames. \textbf{Finally}, building on our improvements in long-term video generation capabilities, we further enhance model diversity by leveraging the greater variability in text data compared to images. We implement joint image-to-video(I2V) and text-to-video(T2V) training based on the DiT~\cite{dit} model, incorporating extensive text augmentation on autonomous driving datasets. Trained on over 1700 hours of data, InfinityDrive can generate a wide range of driving scenarios, ensuring robustness and adaptability across diverse environments.

Overall, our contributions are three-fold: \textbf{1)}We introduce an efficient spatio-temporal co-modeling module with an extended temporal training strategy, enabling the generation of high-resolution (576$\times$1024) video with consistent spatial and temporal coherence. \textbf{2)}We implement memory injection and retention mechanisms combined with a memory curve adaptive loss to prevent cumulative errors and maintain coherence with conditions, enabling consistent, high-quality video generation lasting over 1,500 frames(2 minutes). \textbf{3)}We conduct comprehensive experiments across multiple datasets to evaluate the effectiveness of InfinityDrive. Our model achieves state-of-the-art performance in generating high-fidelity, diverse, and temporally consistent minute-scale videos.

\begin{table*}[ht]
  \centering
 \scalebox{0.7}{
 \begin{tabular}{c|c|c|c|c|c|c|c}
    \hline
     \multirow{2}{*}{Methods}  & \multicolumn{3}{|c|} {Representation Capacity} &  \multicolumn{2}{|c|}{Long-term Rollout} & \multicolumn{2}{|c}{Generalization } \\
    & Frame Rate & Training Frames & Resolution & Duration & Iteration times & Data Scale & Base Model \\
    \hline
    DriveGAN~\cite{drivegan} & 8Hz & - & $256\times256$ &  - & - & 160h & LSTM  \\
    DriveDreamer~\cite{drivedreamer} & 2Hz & 32  & $128\times192$  & 8s  & 1 & 5h & SD\\
    DrivingDiffusion~\cite{drivediffusion} & 2Hz &  6  & $512\times512$  &  4s & 2 & 5h & SD\\
    Drive-WM~\cite{drive-wm}  & 2Hz & 8  & $192\times384$  &  $<20s$ &  5 & 5h & SD\\
    WoVoGen~\cite{wovogen} & 2Hz &  6 & $256\times448$  & 3s  &  1 & 5h &  SD \\
    ADriver-I~\cite{adriver-i} & 2Hz &  - & $256\times512$  &  - & -  & 300h & LLaVA + SD \\
    Panacea~\cite{panacea} & 2Hz & 8 & $256\times512$ & 4s    & 1 & 5h & SD \\
    GAIA-1~\cite{gaia-1} & 25Hz &  - & $288\times512$ &  $> 1 \text{min}$ &  - & 4700h & Transformer \\
    Delphi~\cite{delphi} & 2Hz & 8  & $512\times512$ & 4s & 1 & 5 & SD \\
    DriveScape~\cite{drivescape} & 10Hz & 8  & $576\times1024$  & 4s  & 5  & 5h & SVD \\
    Vista~\cite{vista} & 10Hz & 25  & $576\times1024$  & 15s  & 6  & 1740h & SVD \\
    \hline 
    Ours& 10Hz & \textcolor{red}{128}  & \textcolor{red}{$576\times1024$} & \textcolor{red}{2min}  & \textcolor{red}{12} & 1745h & Transformer  \\
  \end{tabular}
  }
  \caption{Feature-by-feature comparison of driving world models. InfinityDrive can generate long-term driving videos up to 1500 frames with high resolution.}
  \label{tab:comparision of world models}
\end{table*}

 \section{Related Work}
\label{sec:related}

\subsection{World Model in Autonoumous Driving}
World model infers plausible future states of the world based on historical observations and alternative ego actions~\cite{genie,worldmodel2,gameGAN,position,embodied,copilot4d,sutton}. 
For autonomous driving agents, it is imperative to adapt to complex and out-of-distribution cases and world model offers a promising solution~\cite{survey,gaia-1,drive-wm,drivesim}. To benefit autonomous driving agents, world model should produce data of high fidelity and resolution, and as well as diverse scenarios to meet the generalization requirement. Although many autonomous driving world models are proposed to tackle this task, few of them fully solved the aforementioned challenges~\cite{drivedreamer,drivedreamer2,drivedreamer4d,drivescape,drive-wm,wovogen,adriver-i,drivegan,genad,drivediffusion,delphi}. Most existing methods are limited to produce data of low resolution and frame rate, which fails to convey the fine-grained details of the real world~\cite{gaia-1,drivedreamer,drivegan,drive-wm,wovogen,genad}. Moreover, they are only capable of generating videos of short time windows of 8-25 frames, in which the generated videos tend to closely resemble the initial frames, limiting their diversity~\cite{drivedreamer,drivedreamer2,drive-wm,wovogen,adriver-i,drivegan,genad}. GAIA-1~\cite{gaia-1}, trained on a large corpus of driving data, is able to generate long and diverse driving scenes at a high frame rate; however, its low resolution limits its usefulness. Auto-regressive methods such as Vista~\cite{vista} generate longer video of high resolution, but Vista can only predict up to 15 seconds and its visual quality degrades quickly in complex scenarios. Our Infinity-Drive can produce minute-long driving videos of high fidelity, diversity and high resolution.

\subsection{Long Video Generation}
Video generation represents an effective way to understand and predict the world, and progressed remarkably in the recent years thanks to the advancement of diffusion models~\cite{ddpm,diffusion,diffusion2,vdm,dit,video,video2,videomodel} and auto-regression method\cite{videogpt,arlon,artv}. Current video generation methods often build on a pre-trained diffusion models and achieve video generation through further training on video-text paired data~\cite{t2v0,livephoto,videocrafter,video5modelscope,make-a-video,t2i,emu,4consisti2v}. However, these methods are often restricted to producing short and low-resolution videos due to resource constraints and temporal complexity. 
Several recent methods are proposed to extend to long video generation through aggregating chunks of short videos or conditioning on anchor images~\cite{artv,t2v0,mtvg,videodrafter,nuwaxl,arlon,freelong,freenoise,gen-l,streamingT2V,loong,moviedreamer,seine,pixeldance,imagen,i2vgen-xl}. For instance, FreeNoise~\cite{freenoise} and Gen-L~\cite{gen-l} generate long video through combining short video segments through sliding window mechanisms. StreamingT2V~\cite{streamingT2V} utilizes an auto-regressive approach with the injection of anchor frames and long-short memory modules to maintain temporal coherence. Loong~\cite{loong} trains an auto-regressive LLM-based model using a short-to-long training schedule with a loss re-weighting scheme. However, these methods are often restricted to object-centric video generation, and their ability to handle the intricacy of driving scenarios remains questionable. Moreover, most existing video generation methods, unlike driving world models, are not designed to produce videos with explicit controllability~\cite{sparsectrl,control,control2,controlnet,VideoComposer,animateanything}. On the other hand, our model is developed to generate driving scenarios of realistic details and complexity with explicit controllability.
 \section{Method}
\label{sec:method}
In this section, we present our approach to overcoming key limitations in driving world models. First, we introduce an Efficient Spatio-Temporal Co-Modeling(~\ref{sec:Efficient Spatio-Temporal Co-Modeling}) module with dynamic information density adjustment and extended temporal scope training, enhancing spatial detail and temporal coherence over long sequences. To address error accumulation, we implement Memory Injection and Retention Mechanisms with a Memory Curve Adaptive Loss (~\ref{sec:Long-Term Rollouts}).
Finally, joint I2V and T2V training with a text augmentation scheme (~\ref{sec:Enhanced Scenario Diversity}) expands the scenario diversity, enabling our model to adapt to a broad range of conditions. 

\begin{figure*}[t]
  \centering
   \includegraphics[width=1.0\linewidth]{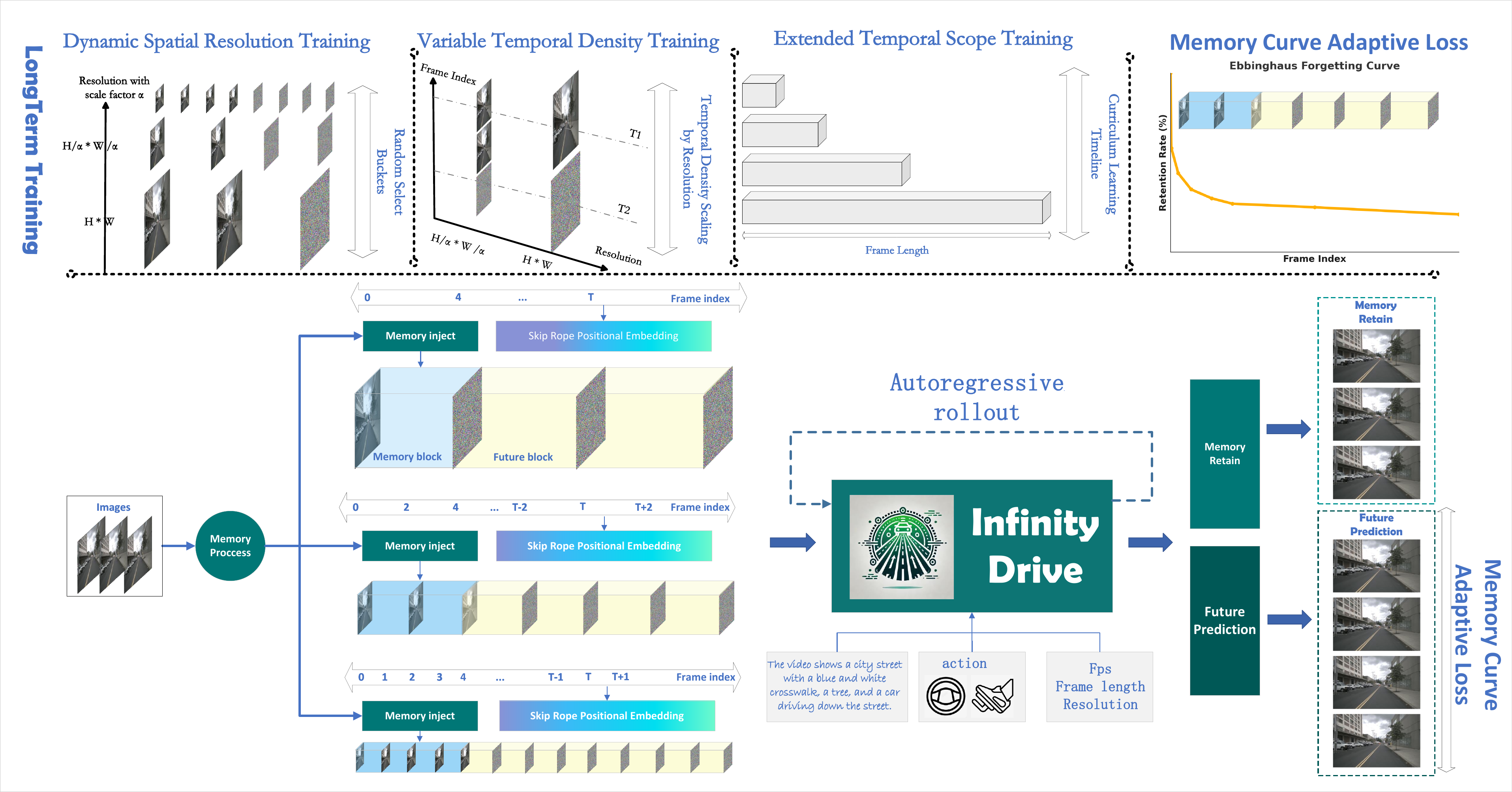}
   \caption{InfinityDrive Pipeline: We introduce an efficient spatio-temporal co-modeling module enhanced with memory injection and retention mechanisms. Combined with long-term training strategies and a memory curve adaptive loss, our model achieves high-resolution video generation lasting over 1500 frames.}
   \label{fig:pipeline}
\end{figure*}


\subsection{Efficient Spatio-Temporal Co-Modeling}
\label{sec:Efficient Spatio-Temporal Co-Modeling}
To address the limitations in spatio-temporal resolution observed in existing driving world models, we propose the \textbf efficient {Spatio-Temporal Co-Modeling}(STCM) module.  Without Spatio-Temporal Co-Modeling (STCM), GPU memory constraints prevent the model from training simultaneously on high resolution (576$\times$1024) and long-term sequences (128 frames). STCM allows the model to prioritize spatial detail at higher resolutions while improving temporal modeling at lower resolutions. 

\subsubsection{Dynamic Information Density Adjustment}
In the standard training configuration, the training data has a base resolution of (H, W), a maximum frame rate of K, and a training sequence length with L frames, enabling observation of L / K seconds of future context. In each training iteration, we first randomly select a resolution scaling factor $\alpha \in [1, 4]$, and then scale the resolution to $(H \times \alpha, W \times \alpha)$.
As a result, the permitted sequence length is proportionally reduced to $L_{curr} = L / (\alpha^2)$. To accommodate the shorter $L_{curr}$, we uniformly sample $L_{curr}$ frames from the original L-frame sequence, while retaining their original indices in the sequence as $P = [..., T - \alpha^2, T, T + \alpha^2,...]$. In addition, for temporal position encoding, we introduce a new positional embedding termed skip-ROPE\cite{rope}, encoding frames at the original K frame rate rather than the downsampled indices. By encoding frames at the original temporal stride, skip-ROPE ensures consistent temporal embeddings across both low and high resolutions. This exposure at lower resolutions allows the model to build a coherent understanding of time, enabling accurate interpretation and prediction of temporal information in high-resolution, long-horizon scenarios that cannot be directly trained due to memory limitations.

Overall, this module dynamically modulates information density, reducing temporal density at higher resolutions to allow the model to focus on critical spatial details without processing dense and redundant temporal information in each frame
Conversely, temporal density is increased at lower resolutions, enabling the model to track changes and maintain coherence over time without incurring significant computational costs. This targeted adjustment supports the generation of high-resolution (1024×576) videos with enhanced temporal consistency.


\subsubsection{Extended Temporal Scope Training Strategy}
To reduce the difficulty of long-term learning, we introduce an extended temporal scope training strategy using a curriculum learning approach. This strategy progressively expands the temporal window, reaching up to 128 frames per iteration, allowing the model to learn extended dependencies and behavioral patterns that are otherwise missed in shorter windows. We start with training on sequences of $L=16$ frames, leveraging the extrapolative capacity of ROPE to use this model as a pretrained foundation. We then extend training to 32, 64 and ultimately 128 frames. In the 16-frame phase, a larger batch size improves training efficiency and stability. As the frame count increases, batch size is reduced accordingly to exceeding the GPU memory limit, but a well-trained, stable pretrained model helps mitigate these limitations.

\subsection{Long-Term Rollouts}
\label{sec:Long-Term Rollouts}
The driving world model performs long-term rollouts by iteratively predicting short-term clips and resetting the condition image from the last clip. However, auto-regressive error accumulation is a common issue in long-duration video generation, where small initial prediction errors amplifies over time, causing significant drift from true conditions. To counter this, we introduce the \textbf{Memory Injection and Retention Mechanisms} along with a \textbf{Memory Curve Adaptive Loss}, that collectively improve consistency in long-term predictions.

\subsubsection{Memory Injection and Retention Mechanisms}

Let input latent\[x = \{ x^1, x^2, \dots, x^T, \dots, x^L \}\] represents the input sequence of frames, where \( L \) is the total number of frames in the sequence. Divide \( x \) into two parts: 1)memory segment \( x_{\text{mem}} = \{ x^1, x^2, \dots, x^M \} \), which consists of the first \( M \) frames. 2)future prediction segment \( x_{\text{future}} = \{ x^{M+1}, \dots, x^L \} \), where \( M < L \).

The choice of \( M \) depends on the resolution. High-resolution inputs have a smaller \( M \), while low-resolution inputs have a larger \( M \) to ensure consistent memory duration across different configurations, providing a stable memory window for the model.

During memory injection, the timestep in the memory module is set to \( t = 0 \), meaning no noise is added, preserving the original historical information. The model only retains this information in the "memory" module without performing any predictions.

The forward process can be expressed by the following formula:

\begin{equation}
 x_t^{T} = \sqrt{\bar{\alpha}_t} x_0^{T} + \sqrt{1 - \bar{\alpha}_t } * \epsilon_t;  \epsilon_t \in \mathcal{N}(0, \mathbb{I});
\end{equation}

\begin{equation}
t = \begin{cases} 0, & \text{if } T \le M \\
\mathcal{U}(0, 1000), & \text{if } T > M 
\end{cases}
\label{eq:add_noise}
\end{equation}

where:\( \bar{\alpha}_t \) are hyperparameters; \( x_t^{T} \) represents the latent at the current frame T at the timestep t; when \( T \le M \), no noise is injected, so the model receives pure historical information in the memory segment.
During future prediction, the memory segment is excluded from the denoising process as its \( t=0 \), while the future segment undergoes the full diffusion denoising process.

\subsubsection{Memory Curve Adaptive Loss}
This loss function is designed to improve memory retention for near-future frames while allowing greater generative freedom for distant-future frames. Frames closer to the memory module prioritize retaining historical details to enhance consistency and reduce error accumulation. In contrast, for frames further into the future, the model is encouraged to rely less on memory and more on its generative capabilities, effectively reducing dependence on past information. 

Inspired by the Ebbinghaus forgetting curve~\cite{ebbinghaus2018gedächtnis}, which describes the gradual decay of memory retention over time, we apply a similar decay principle to the model's loss weights. This approach allows frames near the memory module to have higher loss weights, reinforcing the need for detailed memory retention in these frames. For distant future frames, however, the weights decrease gradually, allowing the model greater freedom to "imagine" or predict independently of memory, thereby reducing memory dependency. The loss can be formulated as:
\begin{equation}
    w(T) = e^{-\lambda T}
  \label{eq:memory_loss} 
\end{equation}

\begin{equation}
L^T_{\text{MSE}} = \mathbb{E}_{t, x_0, \epsilon} \left[ \| \epsilon^T - \epsilon_\theta(x^T_t, t) \|^2 \right]
  \label{eq:mse_loss} 
\end{equation}

\begin{equation}
L^T_{\text{VB}} = \mathbb{E}_{t} \left[ \text{KL}\left( q(x^T_{t-1} | x^T_t, x^T_0) \parallel p_\theta(x^T_{t-1} | x^T_t) \right) \right]
  \label{eq:vb_loss} 
\end{equation}

\begin{equation}
L_{\text{total}} = \sum_{T = 0}^{L} w(T)* (L^T_{\text{MSE}} + L^T_{\text{VB}}) 
 \label{eq:total_loss} 
\end{equation}

where: \( T \) represents the normalized frame index in the sequence, indicating its distance from the memory module. \( \lambda \) is the decay rate parameter, which controls how quickly the weight decreases over time. \( L^T_{\text{VB}} \) is the loss function used in \cite{dit}.


\subsection{Enhanced Scenario Diversity}
\label{sec:Enhanced Scenario Diversity}
Text data introduces greater variability than images, enhancing diversity in training. However, some world models (e.g., SVD~\cite{svd}, Vista~\cite{vista}) lack text-based generation, which limits their diversity. Additionally, diverse text captions are essential for robust video generation. Therefore we develop a scheme to re-caption autonomous driving datasets by leveraging a multi-modal foundation model~\cite{internlm}.
In particular, we curate prompts that take the camera position information as prefix, and in the rest of the prompts we ask the model to describe the scene in the video clip. The prompts are formatted to extract captions describing scenic information such as weather, time, location, and contextual information such as lane layout and objects, and dynamic information such as actions of the ego and other vehicles. 
Examples can be found in the Appendix. 
 \section{Experiments}
\label{sec:exp}
In this section, we first introduce Experimental Setup in ~\ref{sec: setup}, then demonstrate InfinityDrive’s strengths in long-term video generation(~\ref{sec: Long-term Generation}), including fidelity(~\ref{sec: High fidelity Generation}), consistency(~\ref{sec: consistency}), and diversity(~\ref{sec: diversity}). Finally, we conduct ablation studies on our key designs in \ref{sec: ablation study}. For more implementation details and experimental results, please refer to Appendix.

\subsection{Experimental Setup}
\label{sec: setup}
Please refer to Appendix.




\begin{table}[ht]
  \centering
  \scalebox{0.90}{
    \begin{tabular}{c|c|c|c}
      \hline
      Settings & Methods & FID$\downarrow$ & FVD$\downarrow$ \\ \hline
      \multirow{8}{*}{short term} 
      & DriveGAN*~\cite{drivegan} & 73.4 & 502.3 \\
      & DriveDreamer*~\cite{drivedreamer} & 52.6 & 452.0 \\
      & WoVoGen*~\cite{wovogen} & 27.6 & 417.7 \\
      & Drive-WM*~\cite{drive-wm} & 15.8 & 122.7 \\
      & GenAD*~\cite{genad} & 15.4 & 184.0 \\
      & Vista*~\cite{vista} & \textcolor{red}{6.9} & \textcolor{blue}{89.4} \\
      & StreamingT2V~\cite{streamingT2V} & 111.01 & 792.68 \\
      & Ours & \textcolor{blue}{10.93} & \textcolor{red}{70.06} \\ \hline
      \multirow{4}{*}{long term} 
      & SVD-XT~\cite{svd} & 42.99 & 275.08 \\
      & Vista~\cite{vista} & \textcolor{blue}{34.61} & \textcolor{blue}{234.60} \\
      & StreamingT2V-SVDXT~\cite{streamingT2V} & 181.22 & 1255.30 \\
      & Ours & \textcolor{red}{14.92} & \textcolor{red}{113.91} \\ \hline
    \end{tabular}
  }
  \caption{Comparison of generation results; * denotes numbers from the original paper. Short term denotes evaluation with 25 frames and long term denotes 100 frames.}
  \label{tab:comparison}
\end{table}

\subsection{Long-term Generation}
\label{sec: Long-term Generation}
\begin{figure}[t]
  \centering
   \includegraphics[width=1.00\linewidth]{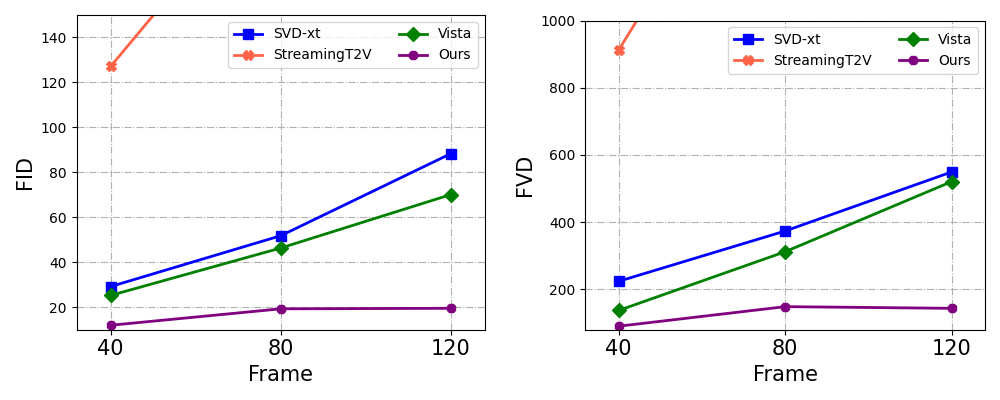}
   \caption{The curves of FID and FVD as world models evolve with duration of time across different frames. We measure FID and FVD at frame 40, 80, 120, using the generated results of previous 40 frames at each time frame point.}
   \label{fig:evolve}
\end{figure}

\textbf{Quantitative Evaluation} Our method demonstrates competitive performance in the short-term setting, achieving the lowest FVD scores, and significantly surpasses other approaches in the long-term setting with lowest FID and FVD. Meanwhile, we visualize the curves of FID and FVD as they evolve with the number of generated frames across different time durations in Fig.~\ref{fig:evolve}. The results reveal that our approach maintains stable FID and FVD scores over long-term durations, achieving consistently high-quality generation across both short and long timeframes. In contrast, methods like Vista, SVD-XT, and Streaming-T2V exhibit a clear upward trend in both FID and FVD scores as the number of frames increases, indicating their lack of ability to ensure consistent generation over long-term durations.

\begin{figure}[t]
  \centering
   \includegraphics[width=1.00\linewidth]{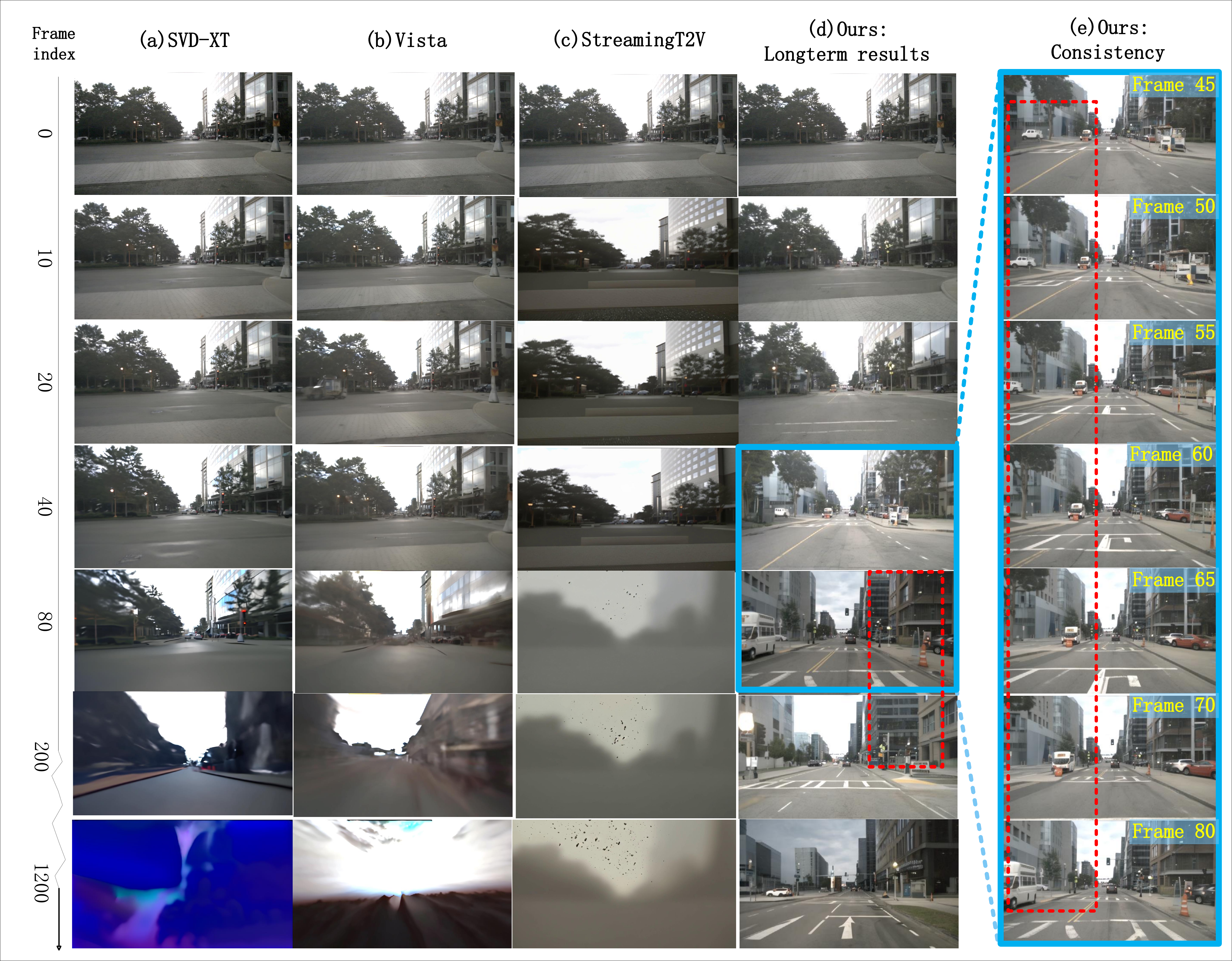}
   \caption{Comparison of long-term video generation results under identical historical image conditions: a) SVD-XT become blur and oversaturated by 80 frames, and eventually fails; b) Vista ultimately lose all details; c) StreamingT2V loses details and displays inconsistencies d) Our model generates up to 1200 frames, preserving spatial detail and maintaining both long- and short-term temporal consistency (highlighted in red boxes).}
   \label{fig:longterm}
\end{figure}

\begin{figure}[t]
  \centering
   \includegraphics[width=1.00\linewidth]{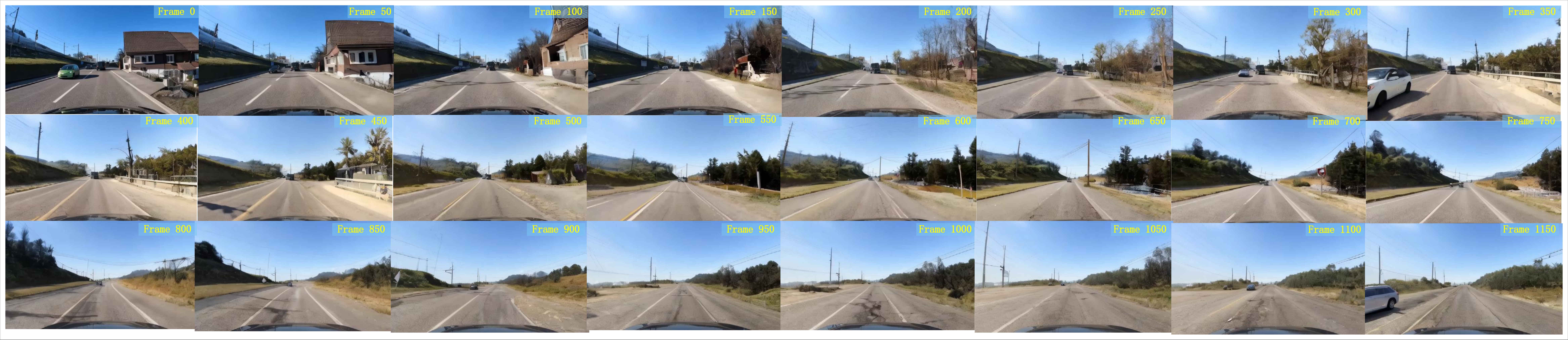}
   \caption{Our long-term generation results on the opendv2k dataset. More results can be found in Appendix.}
   \label{fig:longterm_opendv2k}
\end{figure}

\textbf{Qualitative Evaluation} In our long-term generation task, we use the same history frames as conditions for all methods' rollouts. As shown in Fig.~\ref{fig:longterm}, models such as SVD-XT, Vista, and StreamingT2V fail to maintain visual quality beyond 80 frames (less than 8 seconds), quickly becoming increasingly blurry and eventually losing all discernible details. In contrast, our model consistently delivers high-fidelity results, sustaining quality up to 1,200 frames, approximately 2 minutes. We also present our long-term generated videos, which span up to a two-minute scale, as shown in Fig.~\ref{fig:longterm_ours} and Fig.~\ref{fig:longterm_opendv2k}.

Moreover, even in short-term video generation, previous methods tend to simply repeat historical frames, failing to introduce new contextual elements. Our approach, however, continuously enriches the generated content (Fig.~\ref{fig:longterm} (d) and (e)) with diverse and evolving details, including vehicles approaching from a distance, buildings and parked vehicles in the background, and traffic cones along the road. This demonstrates our model's ability for continuous adaptation and progression rather than merely replicating or extending the initial scene.

\begin{figure}[t]
  \centering
   \includegraphics[width=0.9\linewidth]{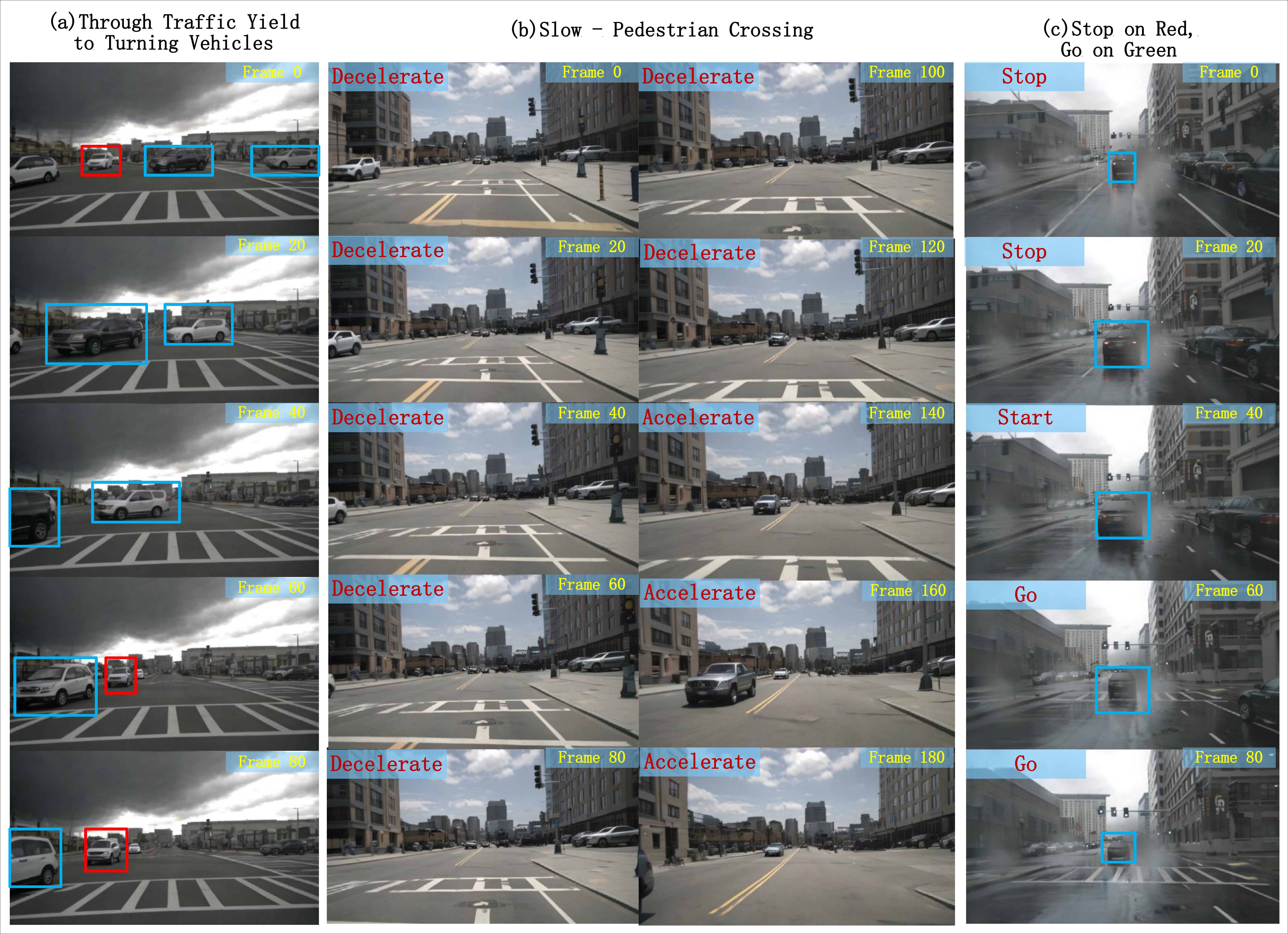}
   \caption{Our model’s long-term prediction capacity captures extended dependencies and behavioral trends, including nuanced vehicle interactions (a), gradual speed adjustments near intersections (b), and adherence to traffic rules (c), which are difficult to learn in shorter time windows.}
   \label{fig:longterm_physic}
\end{figure}

In addition, as illustrated in Fig.~\ref{fig:longterm_physic}, our model’s capacity for long-term prediction enables it to effectively capture and learn extended dependencies and behavioral trends. This includes the nuanced interactions between vehicles (Fig.~\ref{fig:longterm_physic}(a)), gradual speed adjustments, such as deceleration when approaching intersections (Fig.~\ref{fig:longterm_physic}(b)), and the adherence to traffic rules (Fig.~\ref{fig:longterm_physic}(c)), which are challenging to learn within shorter time windows.

\subsection{High fidelity Generation}
\label{sec: High fidelity Generation}
\begin{figure}[t]
  \centering
   \includegraphics[width=1.0\linewidth]{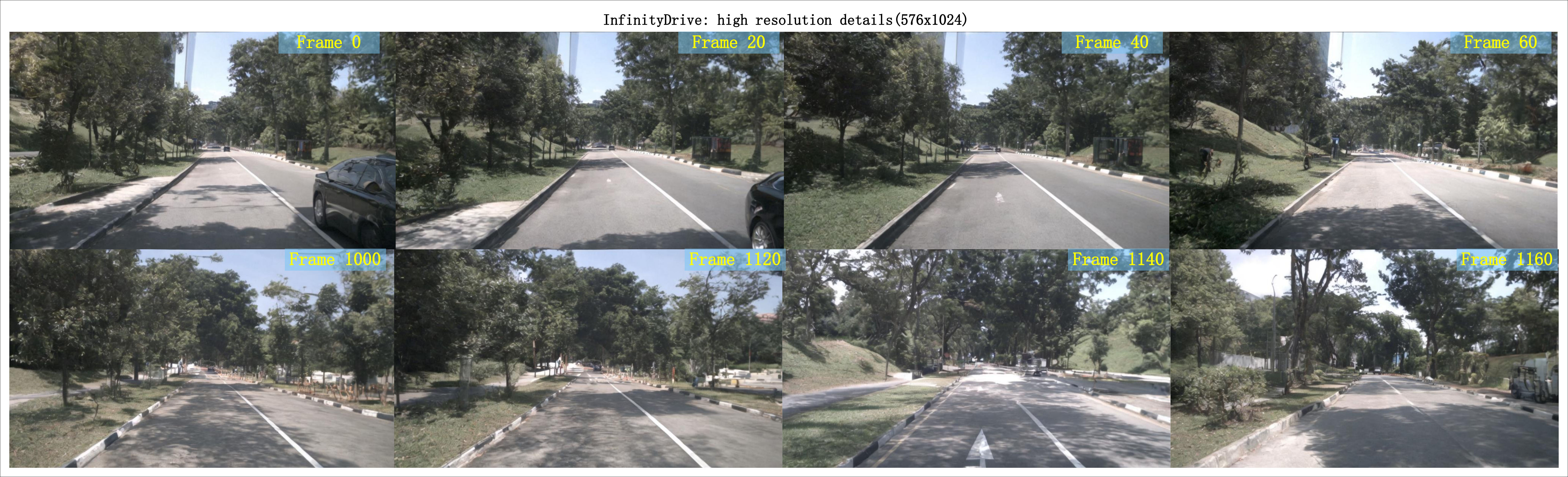}
   \caption{Our high-resolution (576$\times$1024) video generation results.}
   \label{fig:result_hs}
\end{figure}

As shown in Fig.\ref{fig:result_hs}, our high-resolution (576$\times$1024) video generation maintains consistent color saturation and lighting stability across frames, preserving sharp, detailed road textures, vehicle appearances, and background elements throughout.

\subsection{Consistency}
\label{sec: consistency}
Our model achieves coherence in both long-term and short-term generation. As illustrated by the red boxes in the Fig.~\ref{fig:longterm} (d) and (e), buildings remain consistent over longer frames (80–200 frames), while vehicles maintain temporal continuity over shorter frames (45–80 frames). In addition, as shown in Fig.~\ref{fig:longterm_ours}, Even within a minute-level time window, the movement of buildings and vehicles remains consistently synchronized throughout. This ability to produce high-quality, long-duration videos with natural scene progression and temporal consistency underscores the effectiveness of our approach in generating realistic driving scenarios.

\subsection{Diversity }
\label{sec: diversity}
\begin{figure*}[t]
  \centering
   \includegraphics[width=0.9\linewidth]{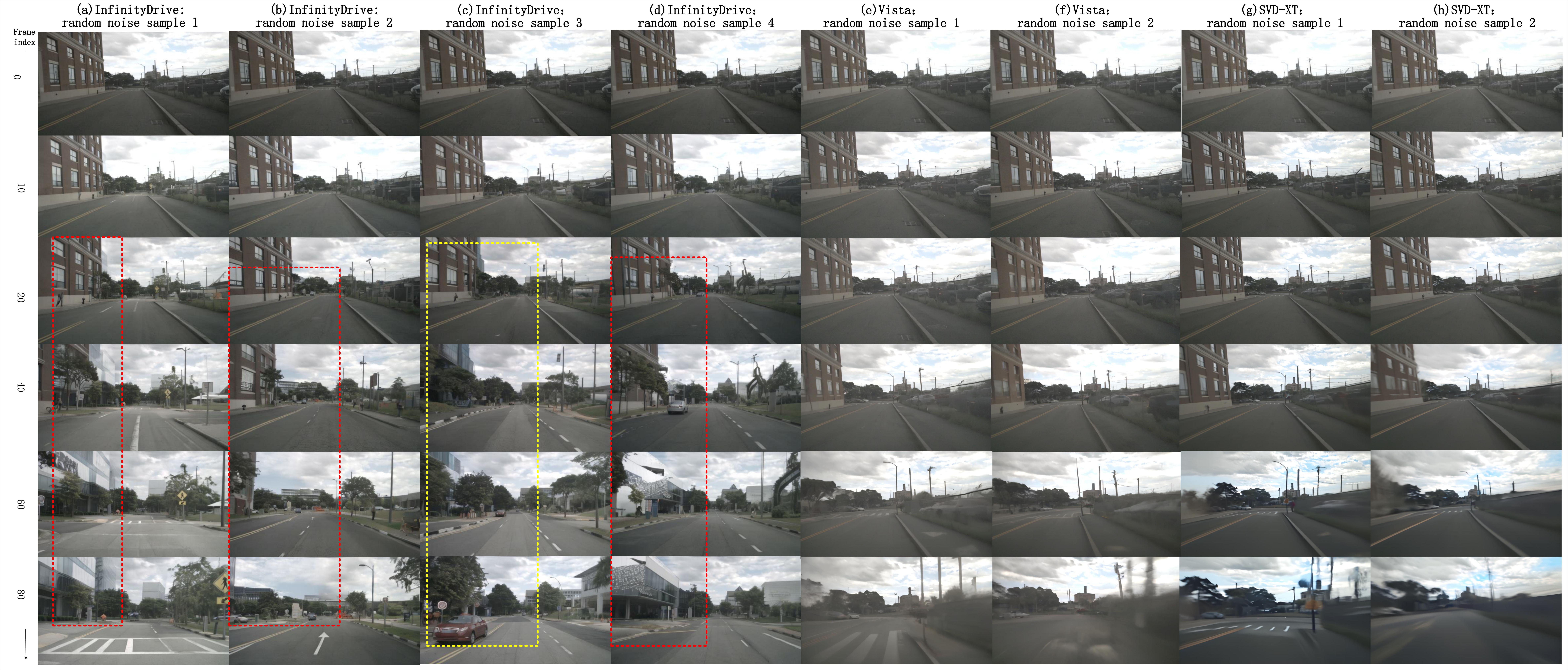}
   \caption{Comparison of diversity in I2V rollouts.}
   \label{fig:diversity}
\end{figure*}

We evaluate the diversity of our models in two settings. Since Vista and SVD are not text-dependent, we assess diversity in the I2V rollouts setting for comparison. Beginning with the same historical frames, we apply different noise inputs for each rollout and visualize the results in Fig.~\ref{fig:diversity}. In every rollout, our model consistently adheres to the initial historical frames(0 - 20 frames in Fig.~\ref{fig:diversity} (a - d)) before evolving uniquely.In addition, as illustrated in Fig.~\ref{fig:diversity}(a-d), our model generates a variety of building structures and diverse traffic participants, introducing new motions and producing variations in both foreground and background elements, while other models remain largely unchanged regardless of the different noise applied.

\begin{figure}[t]
  \centering
   \includegraphics[width=1.0\linewidth]{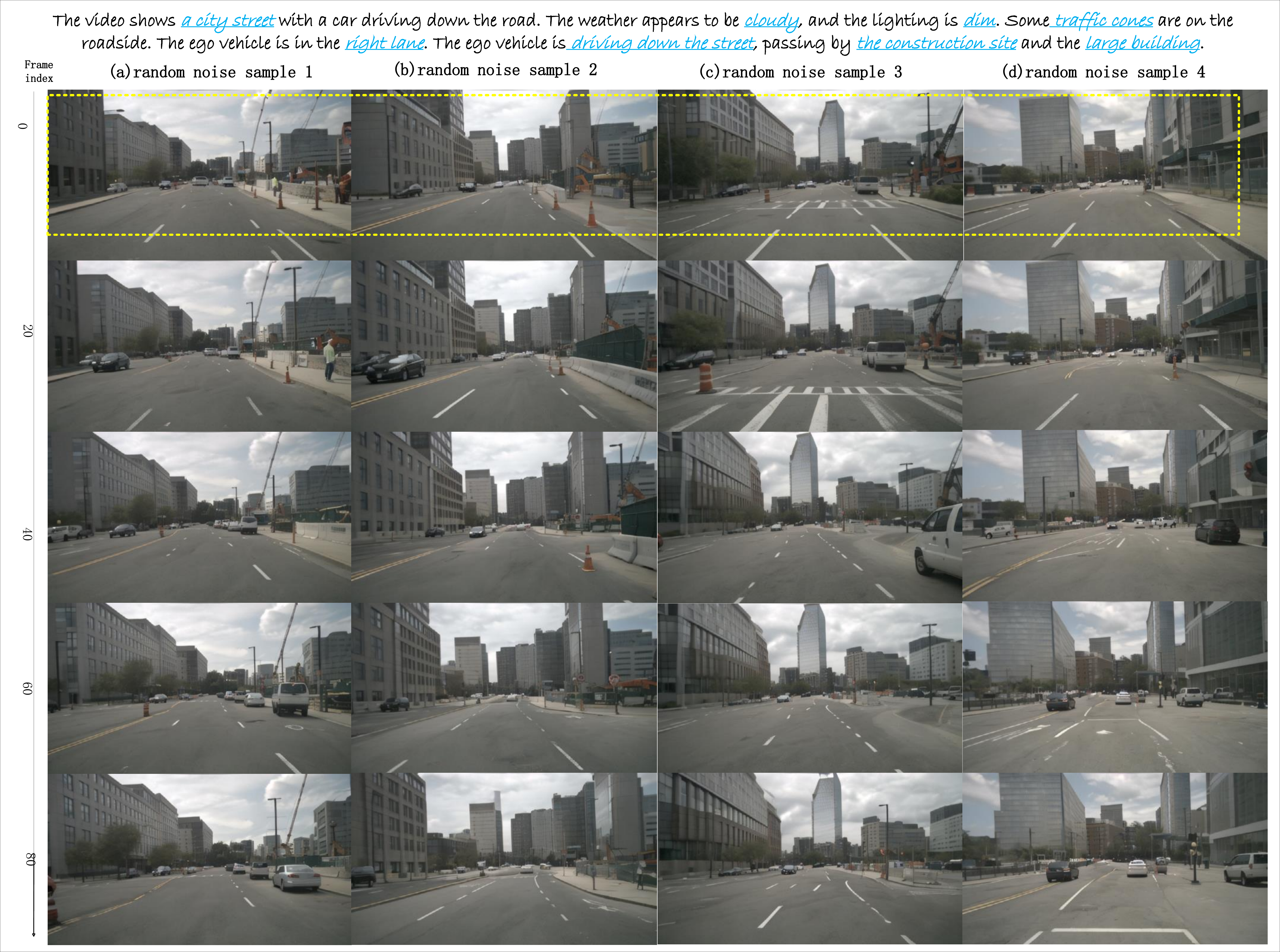}
   \caption{Diversity in T2V rollouts.}
   \label{fig:diversity_t2v}
\end{figure}

Furthermore, using the same text condition with varying noise inputs, as shown in Fig.~\ref{fig:diversity_t2v}, our model consistently adheres to the text prompt, producing identical weather and lighting conditions while generating distinct building structures and diverse traffic participants.


\begin{table}[ht]
  \centering 
 \begin{tabular}{c|c|c|c|c}
 \hline
  \centering
 ETST& STCM  & MCAL & FID$\downarrow$ & FVD$\downarrow$  \\  \hline
    & & & 23.88 & 175.17   \\
  $\checkmark$ & &  &  17.51 & 117.31  \\
  $\checkmark$ &  $\checkmark$ & & 17.40 &  115.62 \\
  $\checkmark$ &  $\checkmark$ &  $\checkmark$ &   14.92 &  113.91\\
\hline
\end{tabular}

\caption{Ablation study results; ETST denotes Extended Temporal Scope Training; STCM denotes Spatio-Temporal Co-Modeling; MCAL denotes Memory Curve Adaptive Loss.}
\label{tab:ablation}
\end{table}

\subsection{Ablation Study}
\label{sec: ablation study}

We validated the effectiveness of the Spatio-Temporal Co-Modeling Module (STCM), Memory Curve Adaptive Loss, and Extended Temporal Scope Training Strategy (ETST). Without these modules, the baseline model achieves an FID of 23.88 and an FVD of 175.17. Incorporating ETST into the training process leads to a significant reduction in FID and FVD, dropping to 17.51 and 117.31 respectively. Adding the STCM module further improves performance, resulting in even lower FID and FVD values. Finally, the inclusion of Memory Curve Adaptive Loss significantly enhances image quality, reducing FID to 14.92 while further decreasing FVD.

\subsubsection{Spatio-Temporal Co-Modeling Module}
\label{sec: Spatio-Temporal Co-Modeling Module}
Without Spatio-Temporal Co-Modeling (STCM), GPU memory limitations prevent the model from training simultaneously on high resolution (576$\times$1024) and long-term sequences (128 frames). STCM enables the model to prioritize spatial detail at high resolutions while enhancing temporal modeling at lower resolutions, enabling the model to learn both detailed spatial features and temporal characteristics.

\begin{figure}[t]
  \centering
   \includegraphics[width=1.0\linewidth]{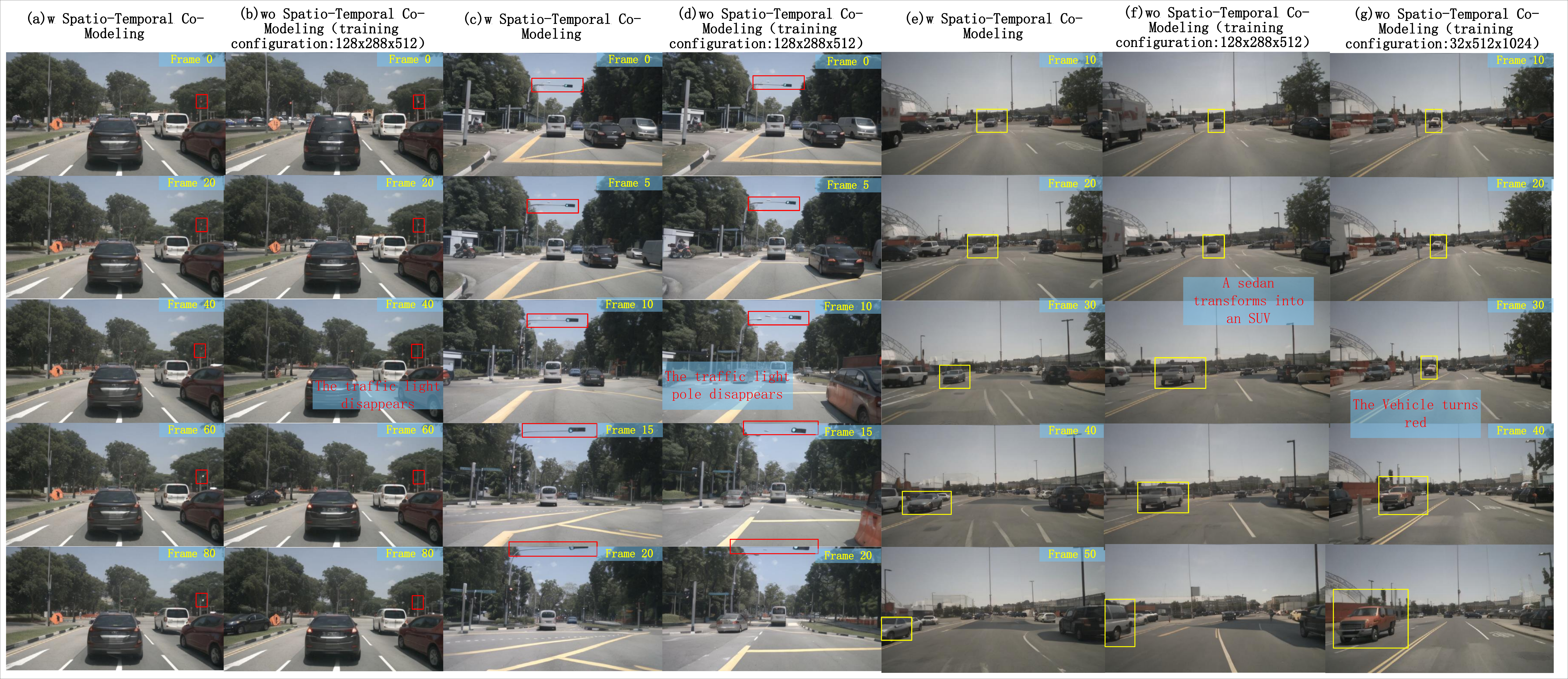}
   \caption{Effectiveness of Spatio-Temporal Co-Modeling (STCM): (a–d) With STCM, our model preserves slender objects details like traffic light and poles (highlighted in red). (e-g) STCM enhances temporal coherence, even for far objects.}
   \label{fig:ablation_comoding}
\end{figure}

As shown in Fig.~\ref{fig:ablation_comoding}, STCM effectively captures spatial detail while enhancing temporal consistency. With STCM, our model better preserves fine details, such as traffic lights and poles (highlighted in the red windows in Fig.~\ref{fig:ablation_comoding} (a-d)), compared to training at single resolution with a 128-frame time window. By reducing the distraction of redundant temporal information, STCM enables the model to focus on learning essential spatial features from high-resolution inputs, improving its accuracy in capturing detailed environmental context. Furthermore, STCM significantly enhances temporal coherence, even for distant objects, as illustrated in Fig.~\ref{fig:ablation_comoding} (e-f). With STCM, the vehicle in the yellow window maintains a consistent appearance across frames. In contrast, without STCM, irregularities arise during both long-term and short-term training (f-g), with the vehicle changing its type and color. This improvement demonstrates the model's ability to leverage increased temporal density at lower resolutions combined with high-resolution learning, ensuring coherence over time even for far objects.

\subsubsection{Memory Curve Adaptive Loss}
\label{sec: memory curve adaptive}
\begin{figure}[t]
  \centering
   \includegraphics[width=0.9\linewidth]{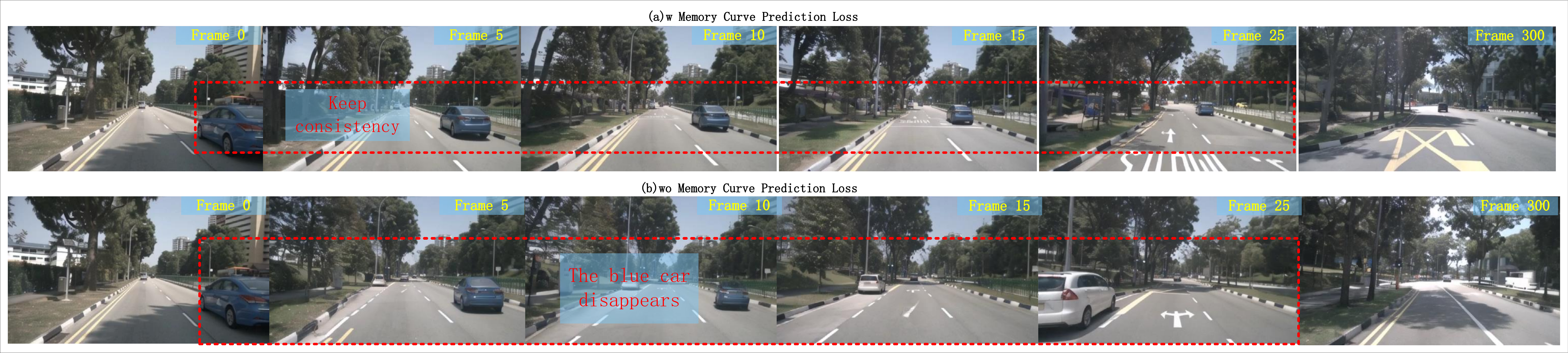}
   \caption{Memory curve adaptive loss helps to preserve coherence with historical states and enhance temporal consistency.}
   \label{fig: memory curve adaptive loss}
\end{figure}

As shown in Fig.~\ref{fig: memory curve adaptive loss}, this loss enables our model to maintain dynamic evolution capabilities, progressively generating new scenarios or states over time while preserving coherence with historical states and enhancing temporal consistency. The red window in Fig.~\ref{fig: memory curve adaptive loss} illustrates this effect, where the blue vehicles remain stable over time.

\section{Conclusion}
We introduce InfinityDrive—the first driving world model, delivering state-of-the-art performance in high fidelity, consistency, and diversity with minute-scale video generation. We conduct comprehensive experiments across multiple datasets to verify the effectiveness of InfinityDrive.  

\noindent \textbf{Limitations and future work} Due to workload constraints, our primary focus is on addressing the challenge of generating temporally consistent long-duration and high fidelity videos. InfinityDrive can serve as a foundational model for a multi-view, controllable driving world model. By generating synthetic driving videos that align with real-world distributions and labeling, such world models provide the richly annotated driving videos crucial for training perception models.

 {
     \small
    \bibliographystyle{ieeenat_fullname}
     \bibliography{main}
 }


\end{document}